# Design and Development of Wall Climbing Robot

Hafiz Muhammad Bilal

# Mechatronics & Control Engineering Department, University of Engineering & Technology
G. T. Road, Lahore 54890 Pakistan

Email: Bilal247@live.com

***Abstract*: Climbing Robots are being developed for applications ranging from cleaning to the inspection of difficult to reach constructions. Climbing robots should be capable of carrying a light payload and climbing on vertical surfaces with ability to cope with obstacles. Regarding adhesion to the surface, they should be able to operate on different surfaces with different adhesion methods to produce strong gripping force using light weight mechanism consuming minimum power. Bearing these facts in mind this paper presents a 4-legged Wall Climbing Robot in which suction power using on board suction pumps is used as an adhesion technique. A Walking gait was developed to provide the robot with a capability for climbing up the wall. The robot's kinematics and motion can be considered as mimicking a technique commonly used in rock-climbing using four limbs to climb. It uses four legs, each with four-degrees-of-freedom (4-DOF) and specially designed suction cups attached to the end of each leg that enable it to manoeuvre itself up the wall and to move in any direction. The end effector can also be replaced with other end effectors designed for different adhesion methods to climb on variety of surfaces.**

## I. Introduction

Climbing robots are very useful devices that can be used in different applications like maintenance, building inspection in construction industries. These systems are usually used in areas where direct access of a human operator is very expensive or very dangerous, due to the presence of an unfriendly environment. Robot that can climb vertically and autonomously along a vertical surface provide considerable military and civilian advantages. When positioned on a high building, the robot, can serve as an observation platform to provide valuable military intelligence and it can assist in search and rescue operations as well. Such a robot can also be used for unmanned clearance of hostile places and can serve as a platform for carrying firearms and explosives. In terms of civilian use, the robot can be used in construction to signal back the status of various operations being implemented at dangerously high levels.

A considerable research has been devoted to these climbing robots and various types of experimental models were proposed so far. The two major issue`s in the design of wall climbing robots are their locomotion and the adhesion methods. With respect to the locomotion type, following types are often considered: the crawler [1], the track wheeled or wheeled [2], [3], the legged [4] types and Omni directional wheel climbing robot [5]. According to the adhesion method, these robots are generally classified into four groups [6]: vacuum or suction cups [7], electrostatic or magnetic [8], dry adhesion [9] and claws for gripping to the rough surface [10]. Recently, new methods for assuring the adhesion, based on biological findings [11], [12] have also been proposed.

There are some limitations attached with almost every adhesion technique and locomotion type. Although, the crawler and wheeled type are fast in movement but it is not adequate to use them in rough environment and also they possess limited ability in crossing cracks and obstacles. Legged climbing robots have the advantages of easily coping with obstacles or cracks found in the environment where they are moving. The adoption of a larger number of limbs can supply redundant support and raises the payload capacity and safety. Whereas the disadvantages are relatively low speed while requiring a complex control.

As far as adhesion techniques are concerned, effectiveness of dry adhesion reduces with continuous operation and pads needs to be replaced. Magnetic adhesion is limited to only metallic surfaces whereas claw like adhesion techniques are effective only in very rough surfaces. Passive suction cups are used to produce adhesion for climbing robot but it is applicable on very smooth surfaces and suction power in passive suction cups is compromised on irregular surfaces.

The climbing robots mostly can be used with a single type of adhesion and little work has been done to develop a robot on which different adhesion methods can be applied to make it able to climb on variety of surfaces.

The proposed design of climbing robot in this paper is a 4-legged robot with the ability to climb and cling on vertical surfaces. Although, active suction through on-board suction pump is used as adhesion technique in the robot presented in this paper but as part of the design goals, it was posited that robot structure should be able to be used with different adhesion techniques. Moreover, the proposed robot design is small, compact and easy-to-carry and is able to independently operate on high walls independent from ground for movement and adhesion. To conduct its missions, the robot can remain statically attached to the wall.

The flow of the paper is that section II presents different aspects of Robot design whereas section III presents testing results and finally, conclusion and possible future work is discussed in the end.

## II. Robot Design and analysis

In this section Mechanical design, pneumatic design, Inverse kinematics and walking and climbing gait design of the robot is discussed.

1. MECHANICAL DESIGN

The robot consists of four legs which are arranged symmetrically around the robot's central body. Each leg has five-degrees-of-freedom (DOF). Four of the DOFs are motorized and the fifth, which is in the gripping device i.e. suction cup assembly mounted on the tip of the leg, is a passive DOF. The first two DOFs, whose axes are perpendicular to the wall, enable the robot to move forward. The two remaining motorized DOFs whose axes are parallel to the wall's plane are designed for determining the distance of the robot from the wall and the angular constraint for the end effector. Using four joints in a leg gives benefit of reduced complexity for controlling the leg movement.

The robot mechanical design is such that other designs of end effector of the legs using different adhesion methods can be used. For example a claw as an end effector figure [6] can also be attached at the tip of each leg to climb on the extreme rough surface like tree or dam concrete wall.

Brackets shown in figure [1-3] are specifically designed for the servo motors used to move the joints. Four actuators per leg were assembled with an end effector's gripping device at the tip of each leg which is a suction cup shown in figure [4-5] specially designed for robot.

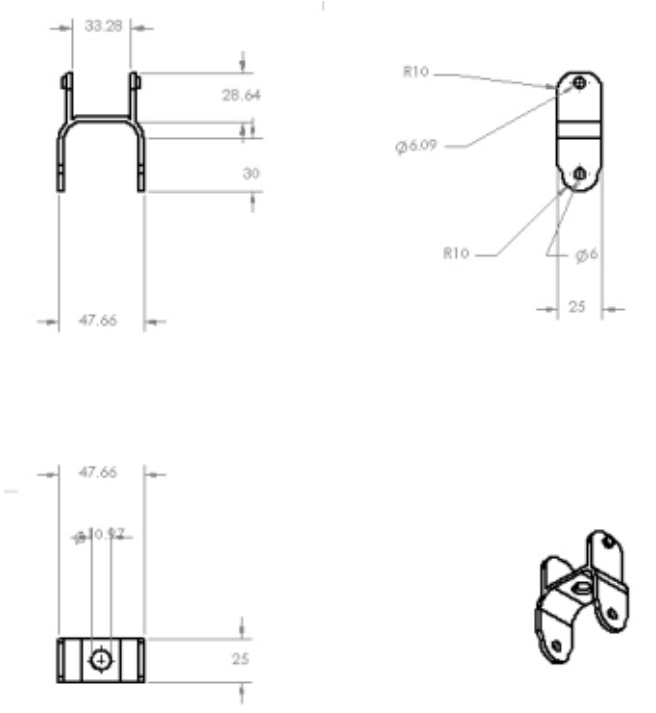

[Figure 1], Motor and end-effector link design

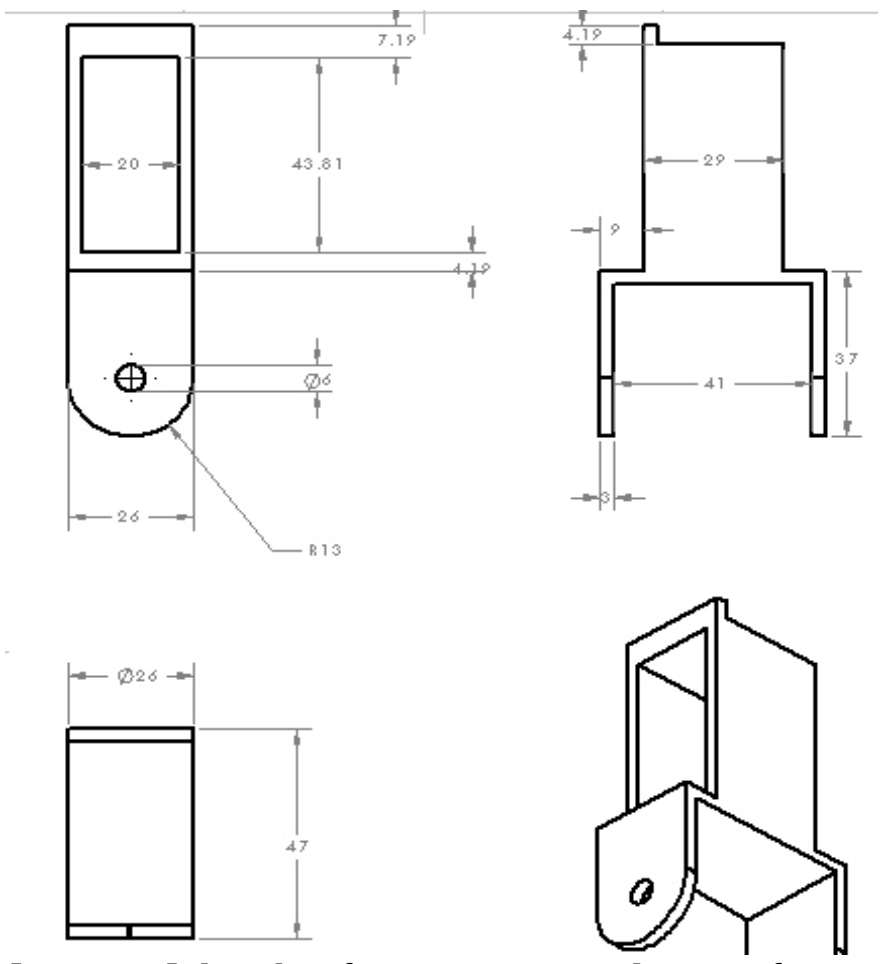

[Figure 4], bracket for motor M1 and M4 in figure [7]

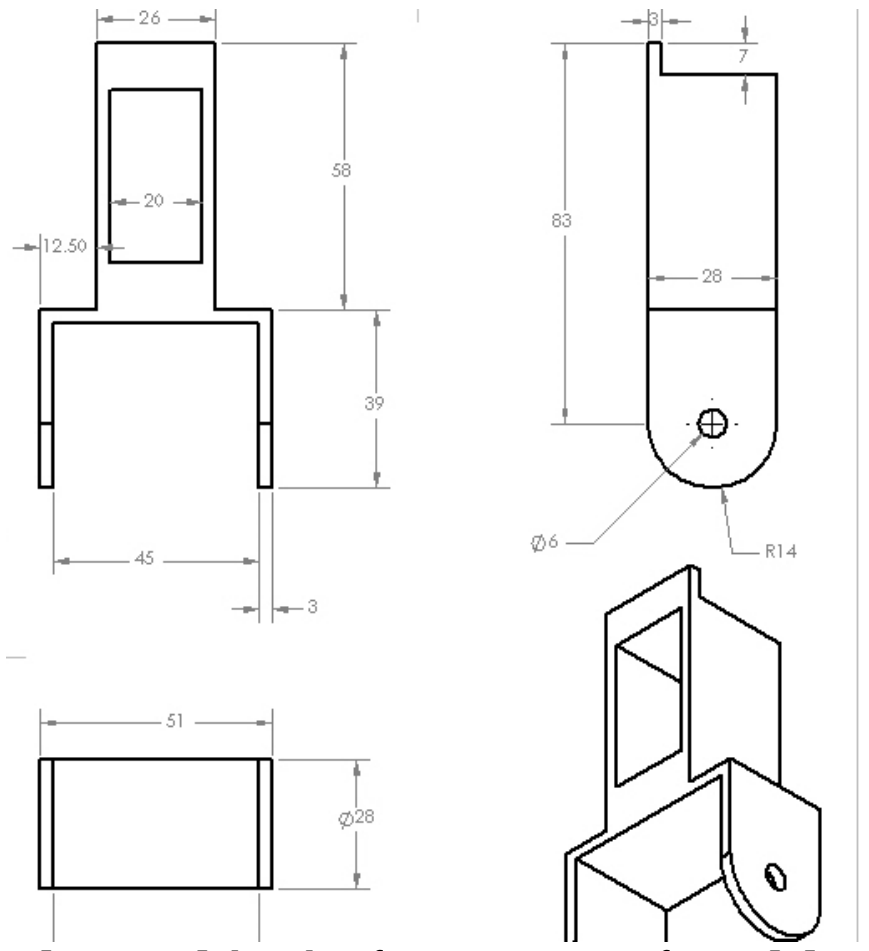

[Figure 5], bracket for motor M3 in figure [7].

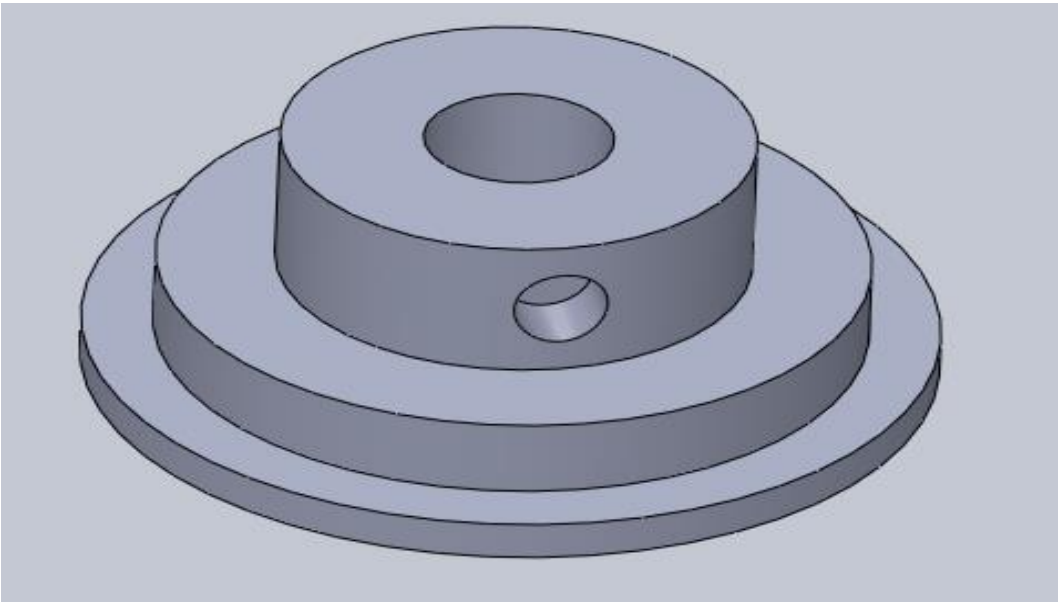

[Figure 4], suction cup support design

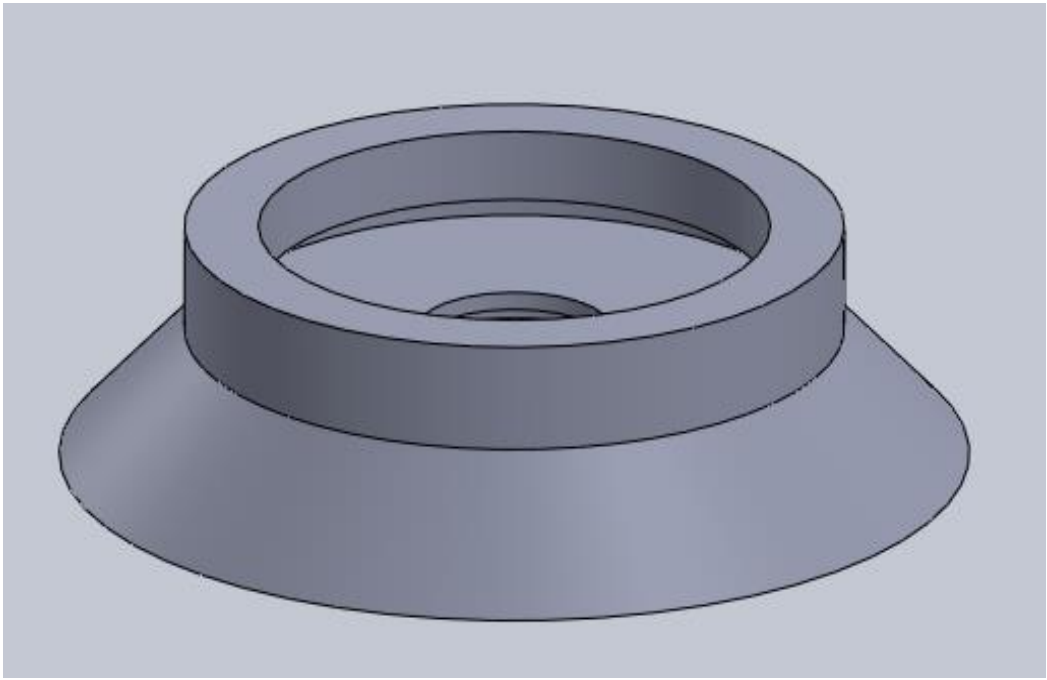

[Figure 5], suction cup design

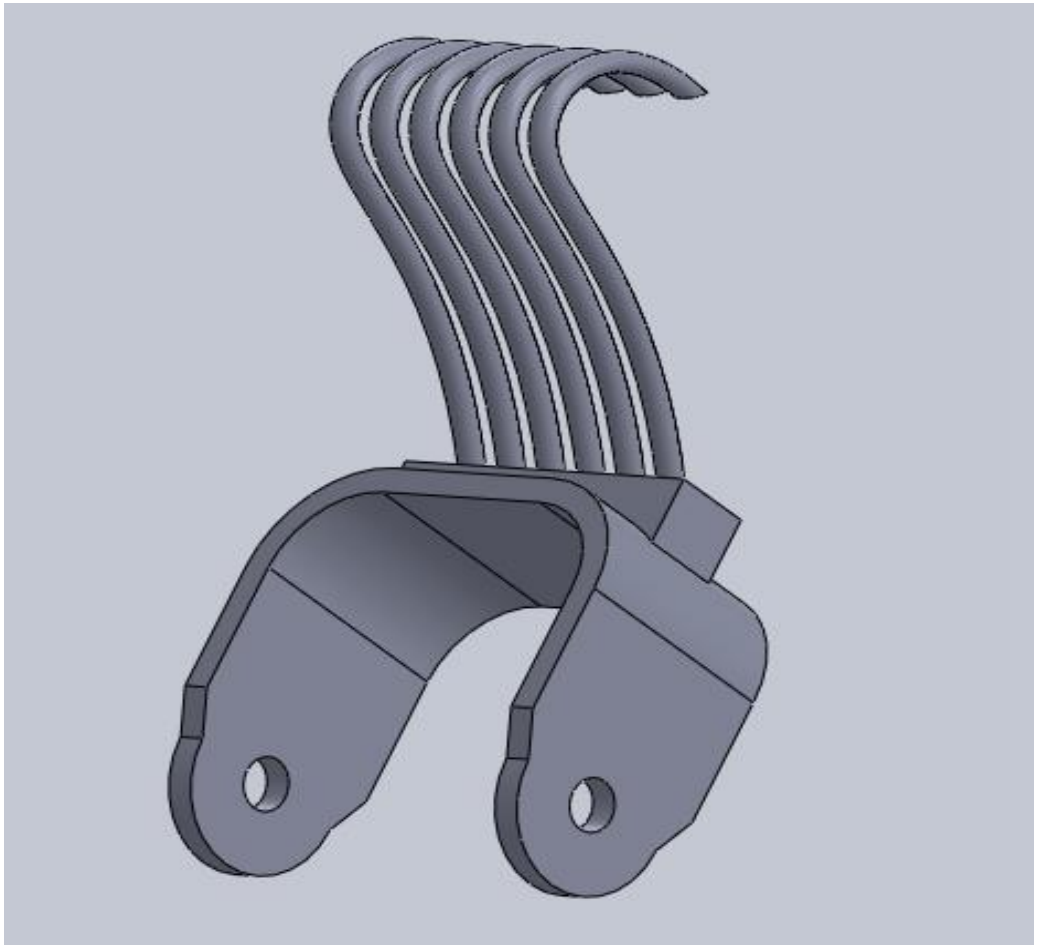

[Figure 6], claw design for extreme rough surfaces

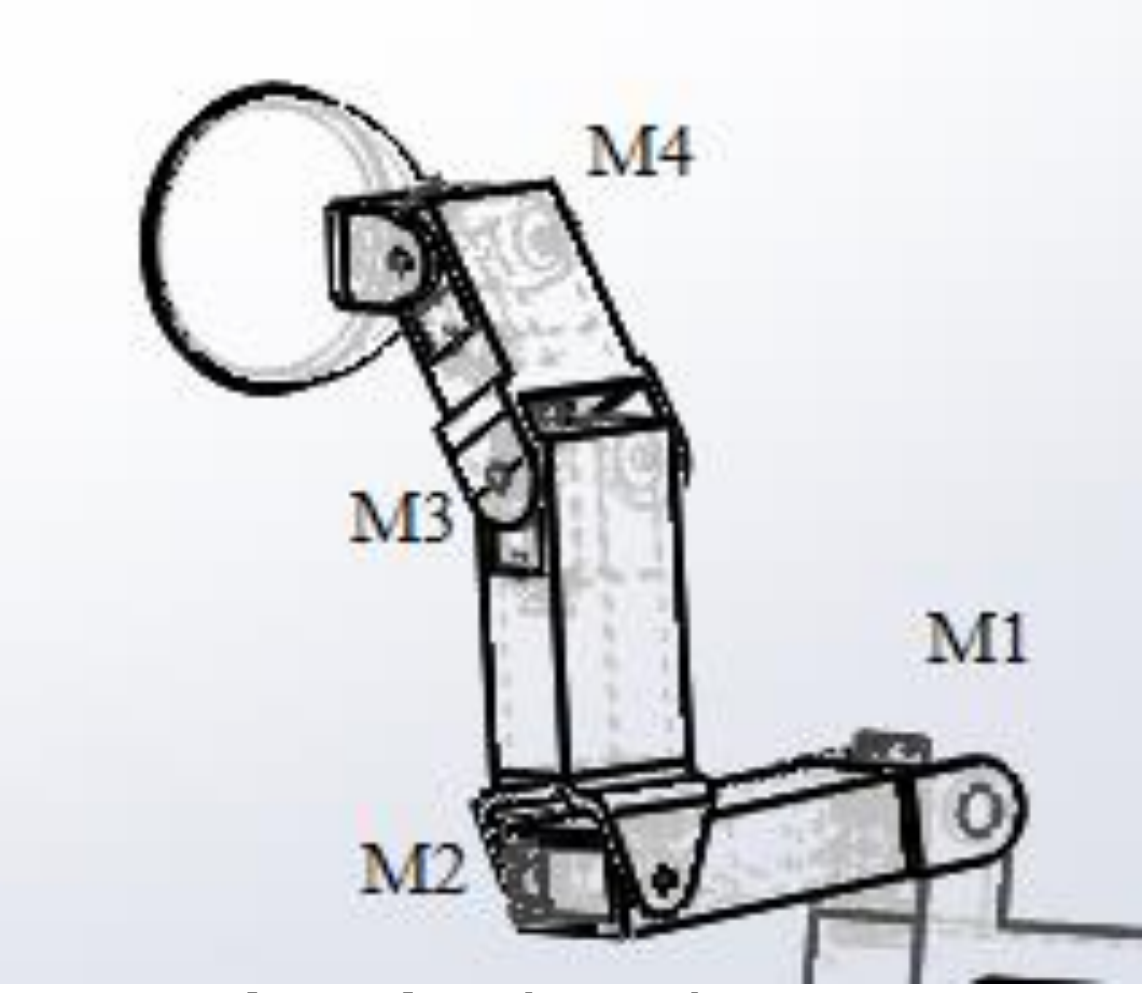

[Figure 7], Leg design with 4 DOF

The design of the leg provides the robot with good gait capability. Furthermore, the robot can change its distance from the wall by extending its legs, to lower or raise itself in relation to the wall's surface according to the surface conditions. After the attachment of the suction cup and upon determination of the distance from the wall by motors 3 and 4 of every leg, the first two motors in each leg drive the robot's movement. Consequently, this leg design has the advantage of decoupling motion in plane (parallel to the wall) and normal to the plane. The complete robot model is shown in figure 8.

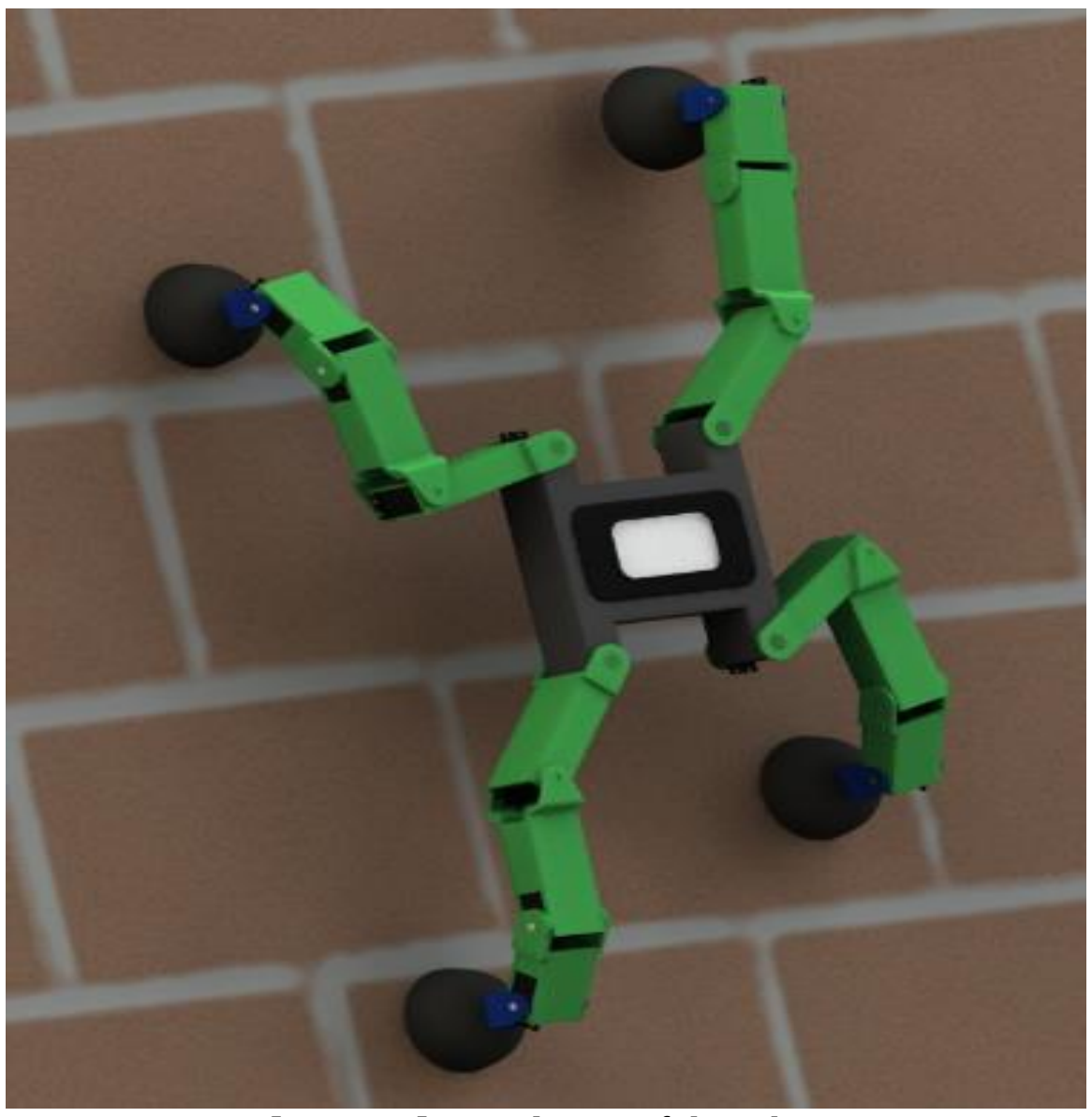
[Figure 8], 3-D design of the robot

2. INVERSE KINEMATICS

Inverse kinematics is used to compute the joint angles for a required end effector's position in xyz plane. As one leg has four joints so there are four angles required to reach at the desired position in xyz space. As mentioned earlier that Robot's leg is designed such that it has the advantage of decoupling the motion in planes parallel to the wall and normal to the wall. Joints denoted by M1 and M2 [figure 7] which are closer to the central base are responsible for the location of the contact point in the xy plane. Whereas other two joints denoted by M3 and M4 [figure 7] are responsible for the distance from the wall and the approach angle of the cup in zy plane. The distance of the central body from the wall is constrained to the defined value Z. With these constraints and assumptions, we need to compute four joint angles to achieve a desired position of the suction cup at the end of a leg.

Two joint angles responsible for contact point of suction cup in xy-plane can be considered as two link manipulator in xy-plane shown in [figure 9].

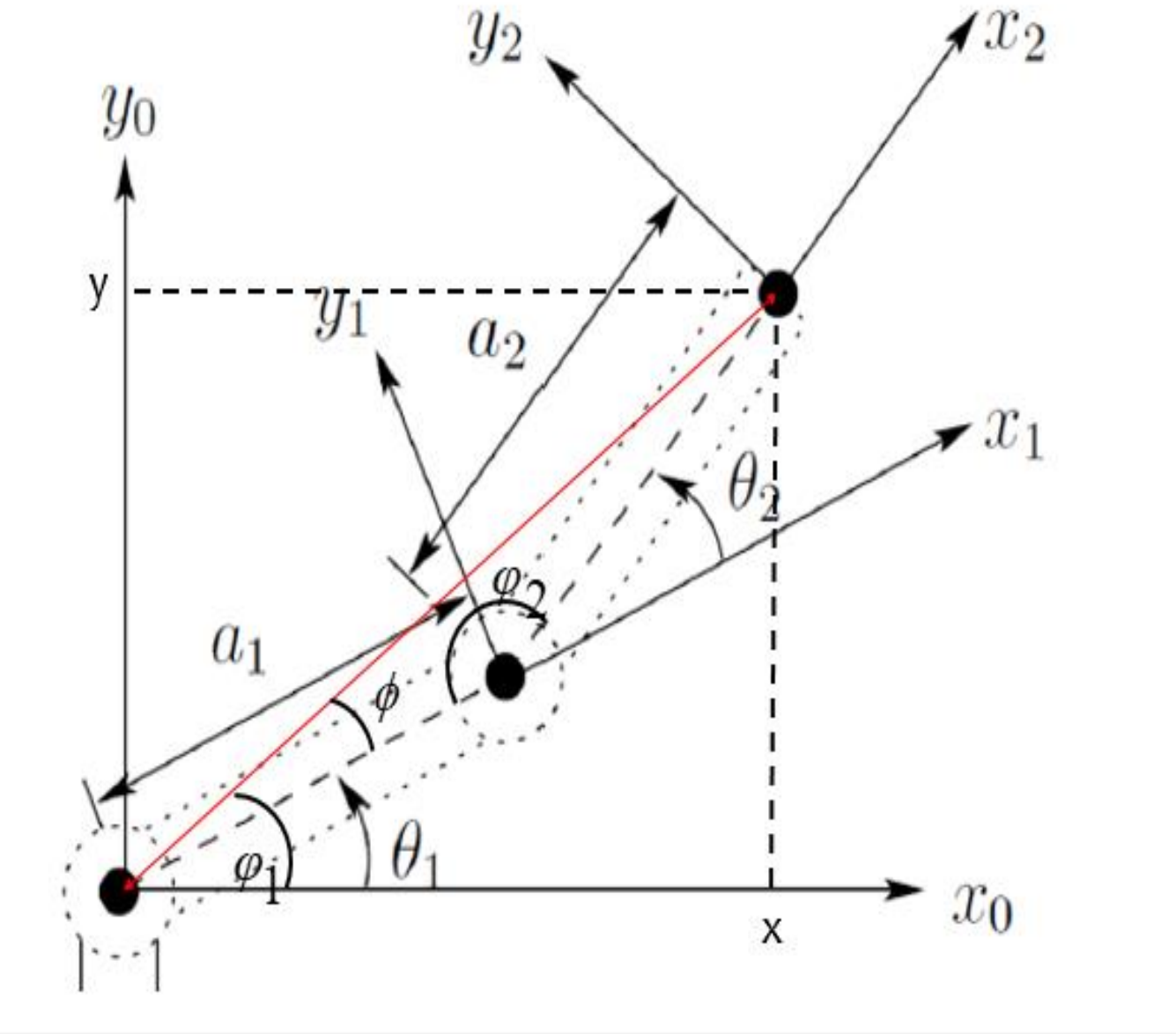

[Figure 9], Inverse kinematcis diagram for two joints angles responsible of forward movement of the robot.

$$\theta_1 = \varphi_1 - \phi$$

$$\theta_1 = \tan^{-1}\left(\frac{y}{x}\right) - \tan^{-1}\left(\frac{a_2 \sin\theta_2}{a_1 + a_2 \cos\theta_2}\right) \rightarrow (1)$$

From law of cosines

$$\left(\sqrt{x^2+y^2}\right)^2 = a_1^2 + a_2^2 - 2a_1 a_2 \cos\varphi_2$$

$$-\cos\varphi_2 = \frac{x^2+y^2-a_1^2-a_2^2}{2a_1 a_2} = D$$

$$180 - \varphi_2 = D \qquad \because \; 180^o - \alpha = -\cos\alpha$$

$$\cos\theta_2 = D \qquad \because \; -\cos\alpha = \cos\beta$$

$$\sin\theta_2 = \pm\sqrt{1-D^2} \qquad \because \; \sin^2\alpha + \cos^2\beta = 1$$

$$\theta_2 = \tan^{-1}\frac{\pm\sqrt{1-D^2}}{(D)} \rightarrow (2) \qquad \because \; \tan\alpha = \frac{\sin\alpha}{\cos\alpha}$$

Two joints M3 &M4 responsible for the distance of robot from the wall can also be considered as two link manipulator in zy-plane shown in [figure 10]. When the leg is attached to the wall the distance Z remain constant.

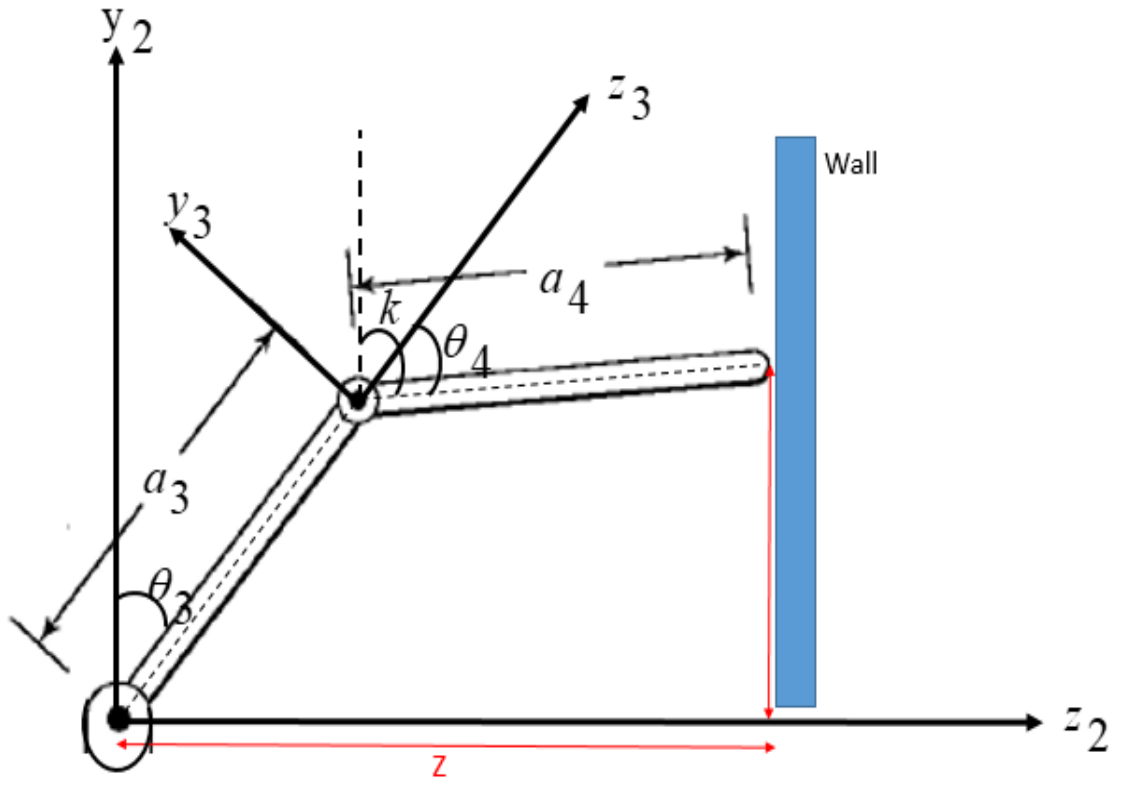

[Figure 10], Inverse kinematcis diagram for two joints angles responsible of maintenaining distance from the wall

$$Z = a_3 \sin\theta_3 + a_4 \sin(k)$$

$$\theta_3 = \sin^{-1}\left(\frac{Z - a_4 \sin(k)}{a_3}\right) \rightarrow (3)$$

$$k = \theta_3 + \theta_4$$

$$\theta_4 = k - \theta_3 \rightarrow (4)$$

## 3. PNEUMATIC DESIGN

Two light weight suction pumps are used for producing suction power in suction cups at the end of each leg. To produce the suction power required to adhere to the wall, on board suction pumps uses diaphragm mechanism to produce suction with the help of a 12 volt dc motor.

The suction power in suction cup at each leg is controlled with solenoid valves operated by a relay board which are actuated by a microcontroller. Figure [11] shows the pneumatic circuit diagram designed for the robot.

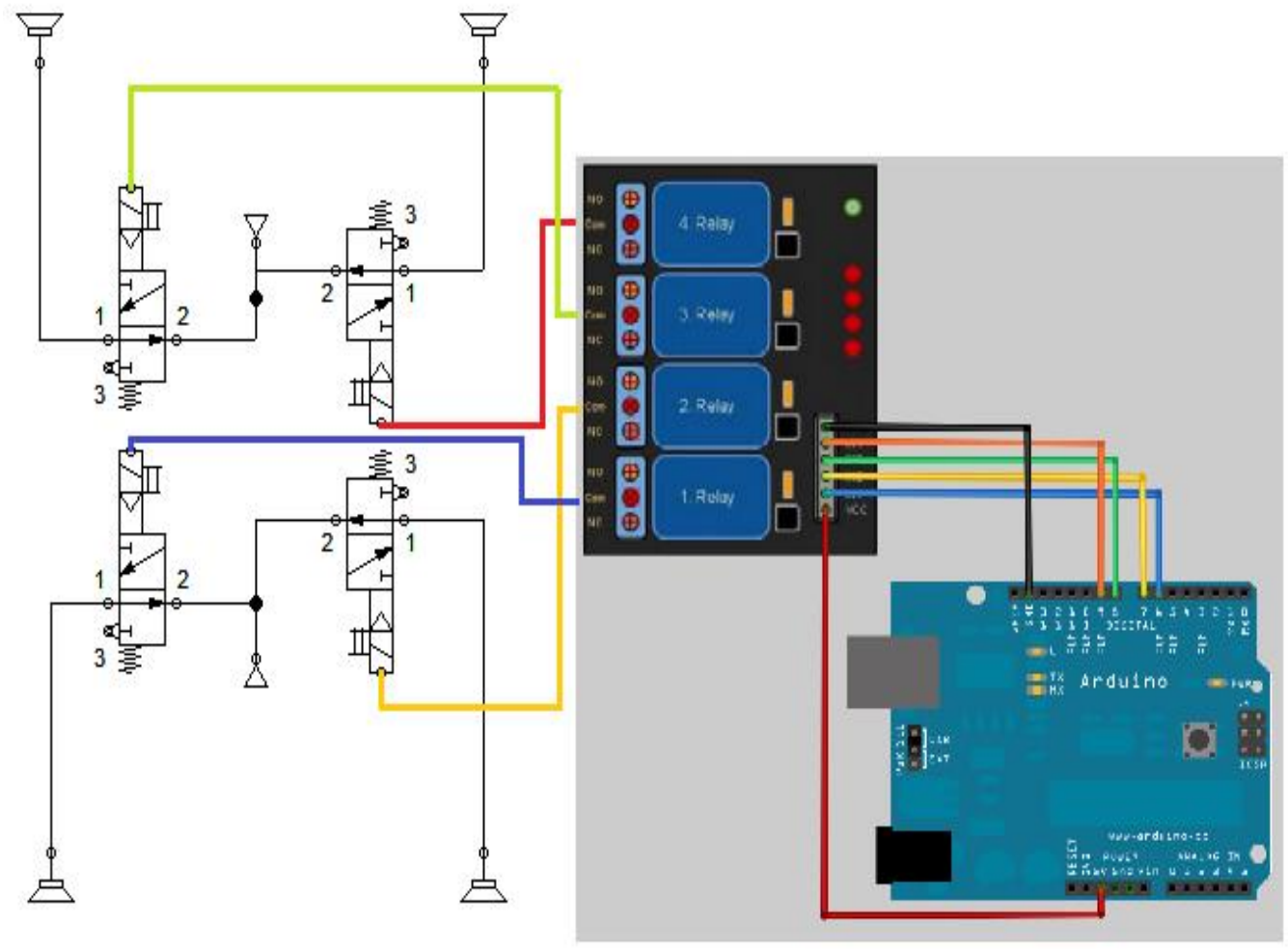

[Figure 11], Pneumatic circuit diagram of the robot for adhesion

## 4. GAIT DESIGN

There are many walking gaits have been designed for legged robots like diagonal gait, crawl gait and creep gait. The gait designed for this robot mimics a human mountain or wall climber. A human climbing on the wall with four limbs climbs in such a way that at any time three of its limbs remain attached to the wall and push the body in upward direction and fourth limb remains in the air moving to next possible point to hold.

Similarly, in this robot three legs remain firmly attached to the wall with the help of suction power in the suction cups and move the robot body in upward direction and in the meantime fourth leg not attached to the wall moves forward to reach to new upward position.

The gait designed for the robot can be explained step by step. When robot is powered up leg 1 is at position $p_1$, leg 2 is at position $p_2$, leg 3 is at position $p_3$ and leg 4 is at position $p_4$ and all legs are attached to the wall with the help of suction cup shown in figure [12-b]. All four legs moves in following four steps repeatedly in order to move the robot upward on the wall.

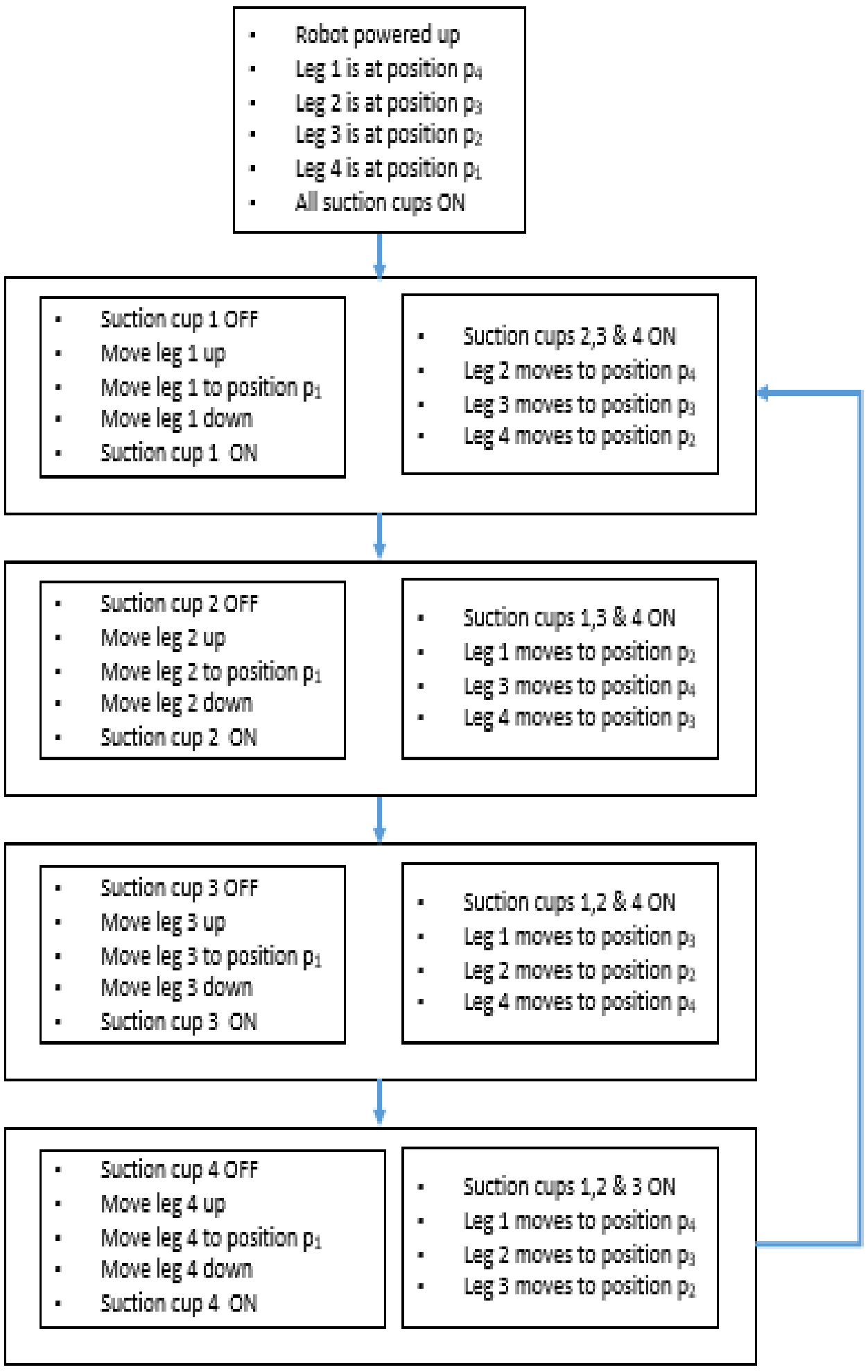

[Figure 12-a], Gait design of the robot

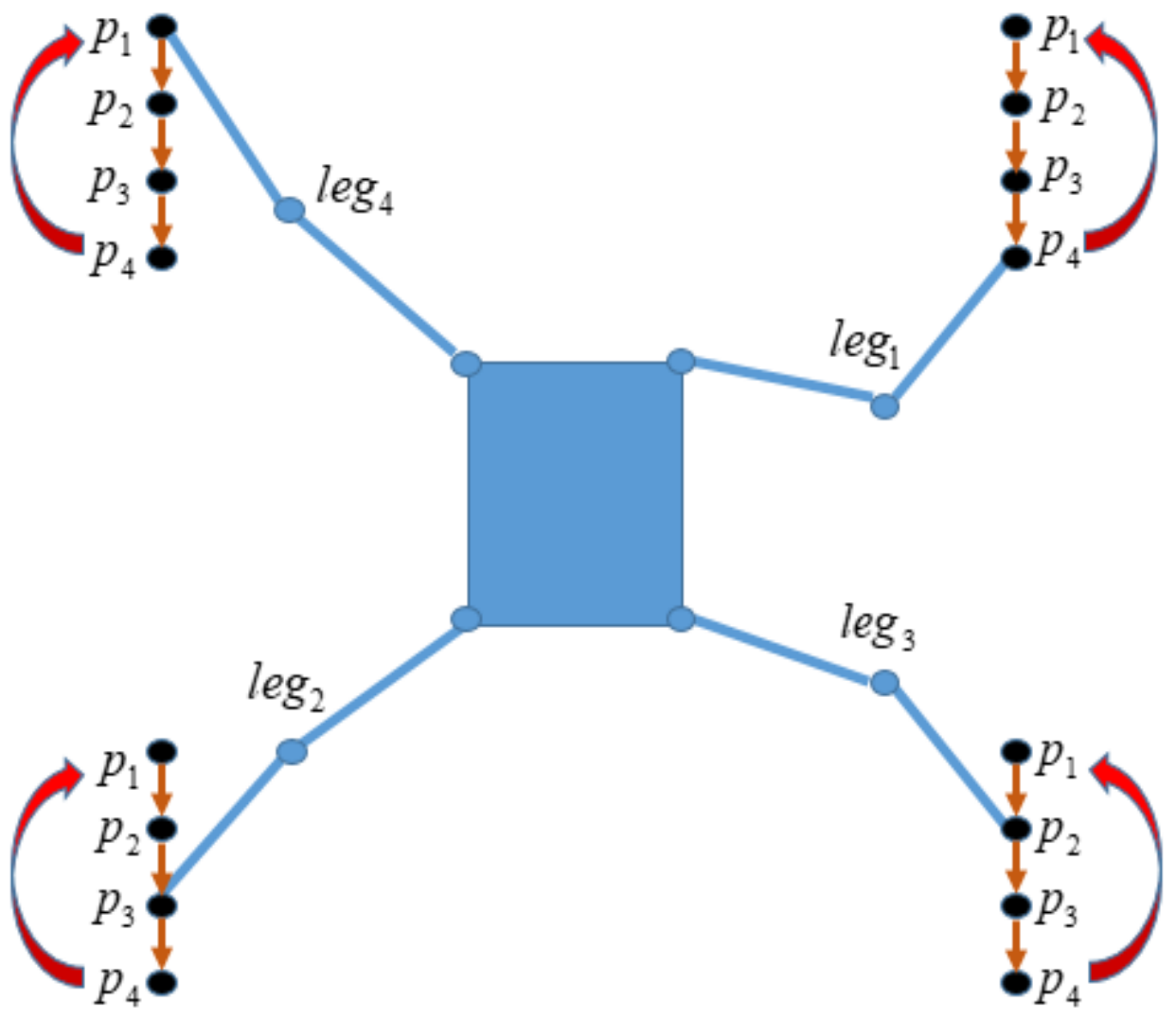

[Figure 12-b], Gait design of the robot

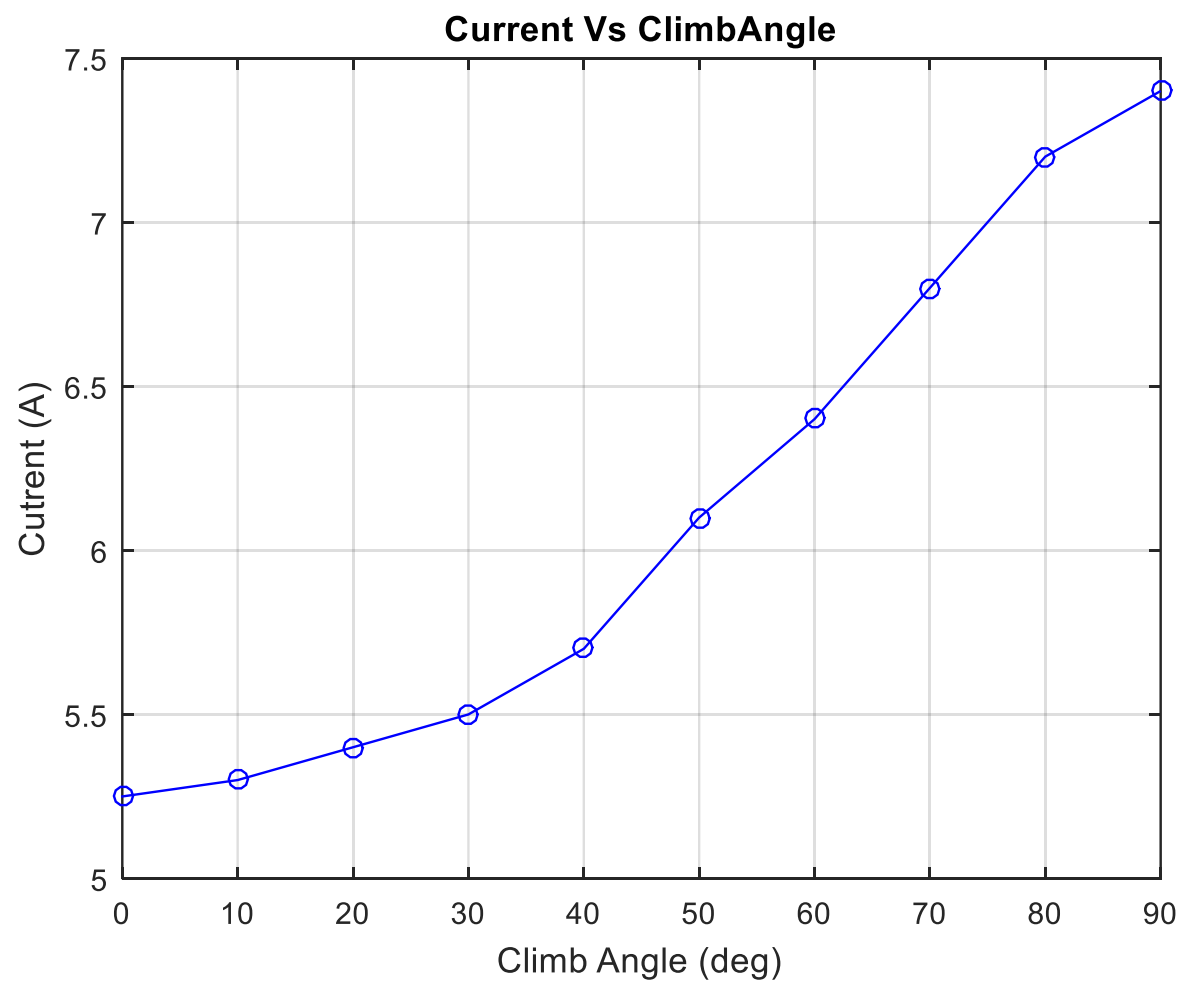

[Figure 14], power consumption vs. climb angle of the robot

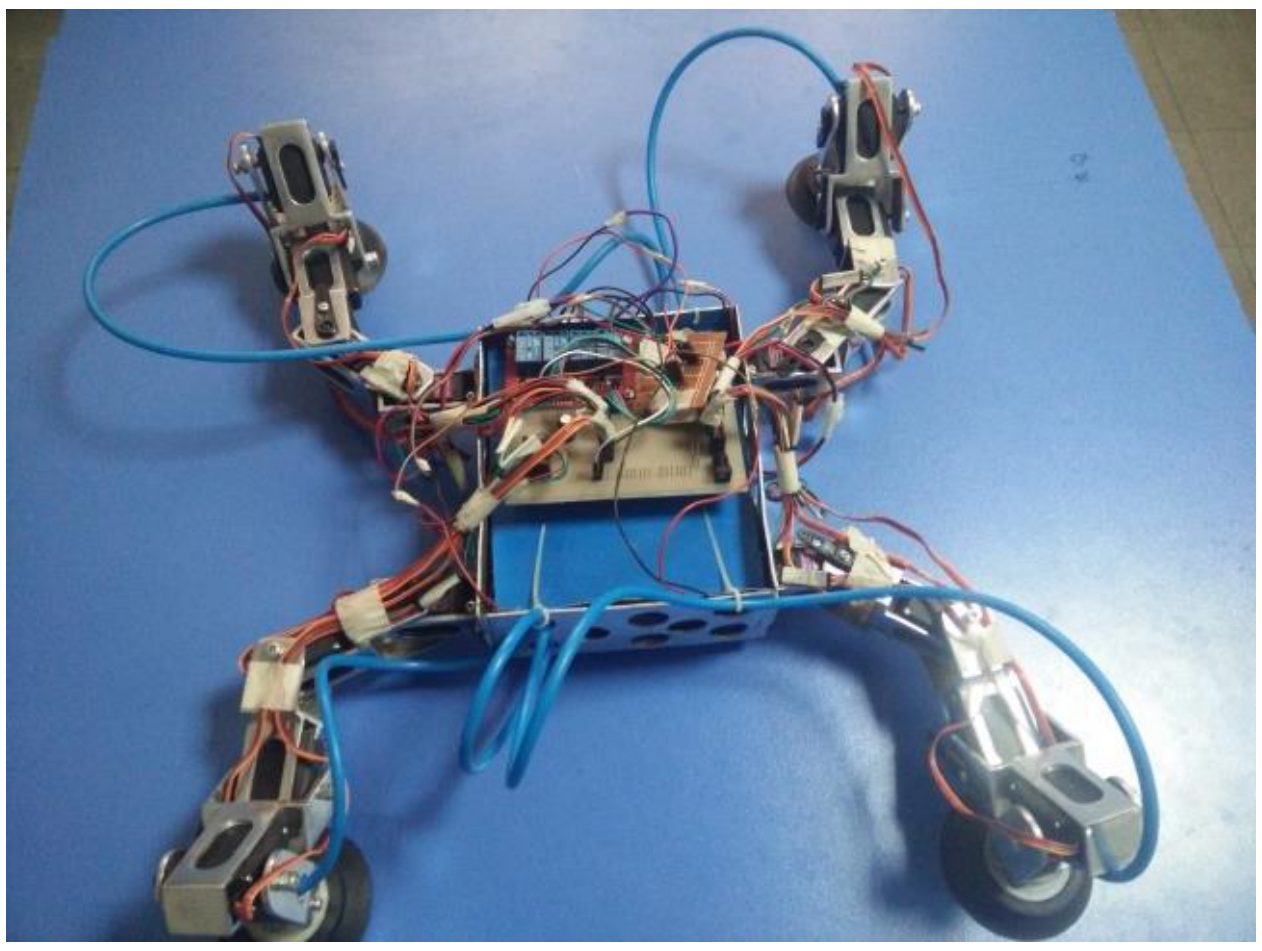

[Figure 15], Robot testing

## III. Testing

The four joint angles required for leg movement were computed offline using equation 1-4 for each leg for position p1, p2, p3 and p4. The designed gait was implemented in aurduino microcontroller. A testing platform was developed to test the Robot's performance. Different inclination angles can be given to the testing platform to test the Robot's performance at different climb angles. Speed of the Robot and power consumption at different climb angles is shown in figure [13] & [14]. Due to onboard power supply and on board suction pump the weight of the robot is heavier, therefore, speed of the robot decreases as the climb angle increases due to slippage of suction cups. Similarly, the power consumption increases as the climbing angle increases due to increased load on suction pumps and servo motor to lift the robot. Figure [15] and figure [16] shows the actual robot in action on the testing platform.

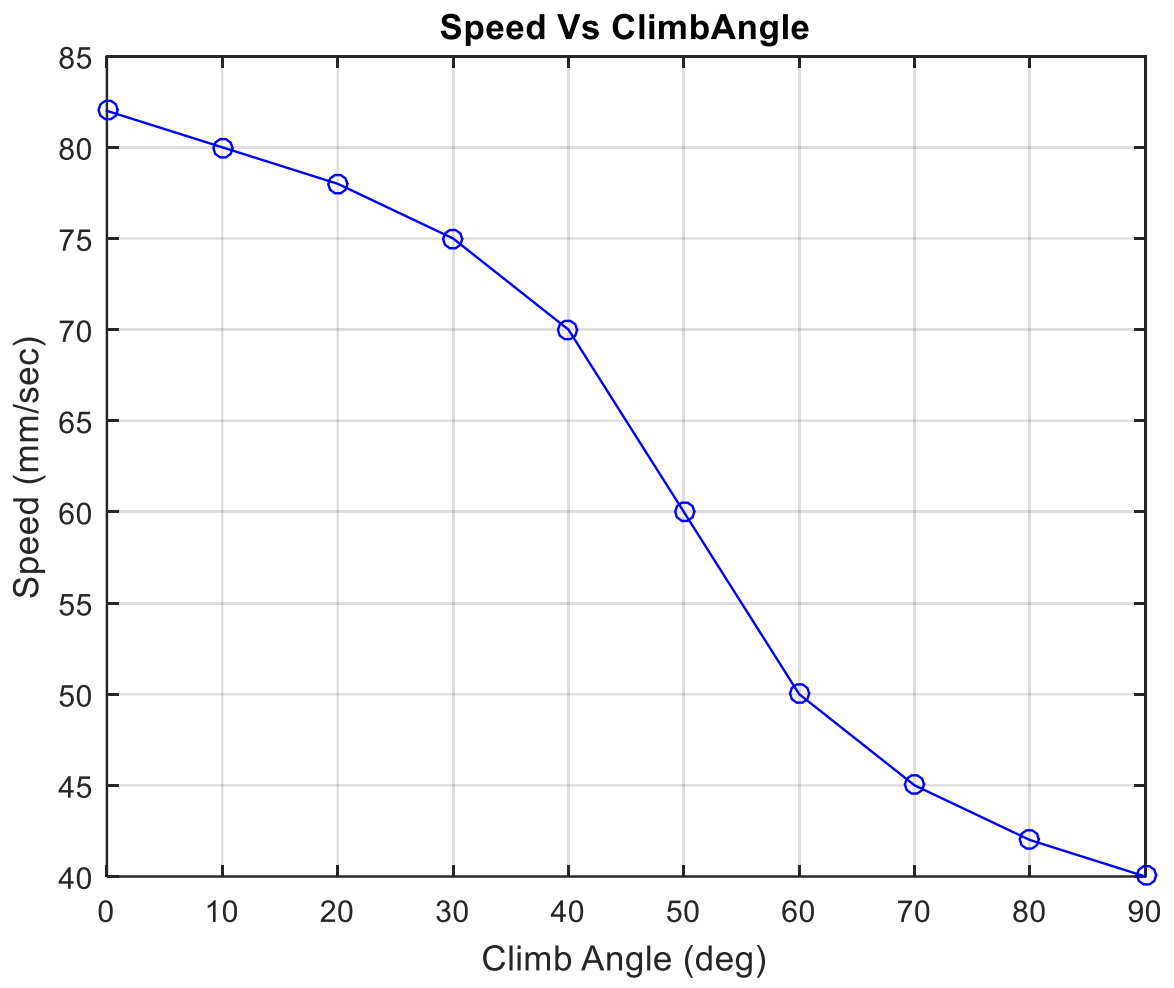

[Figure 13], Speed of the robot vs. climbing angle of the robot

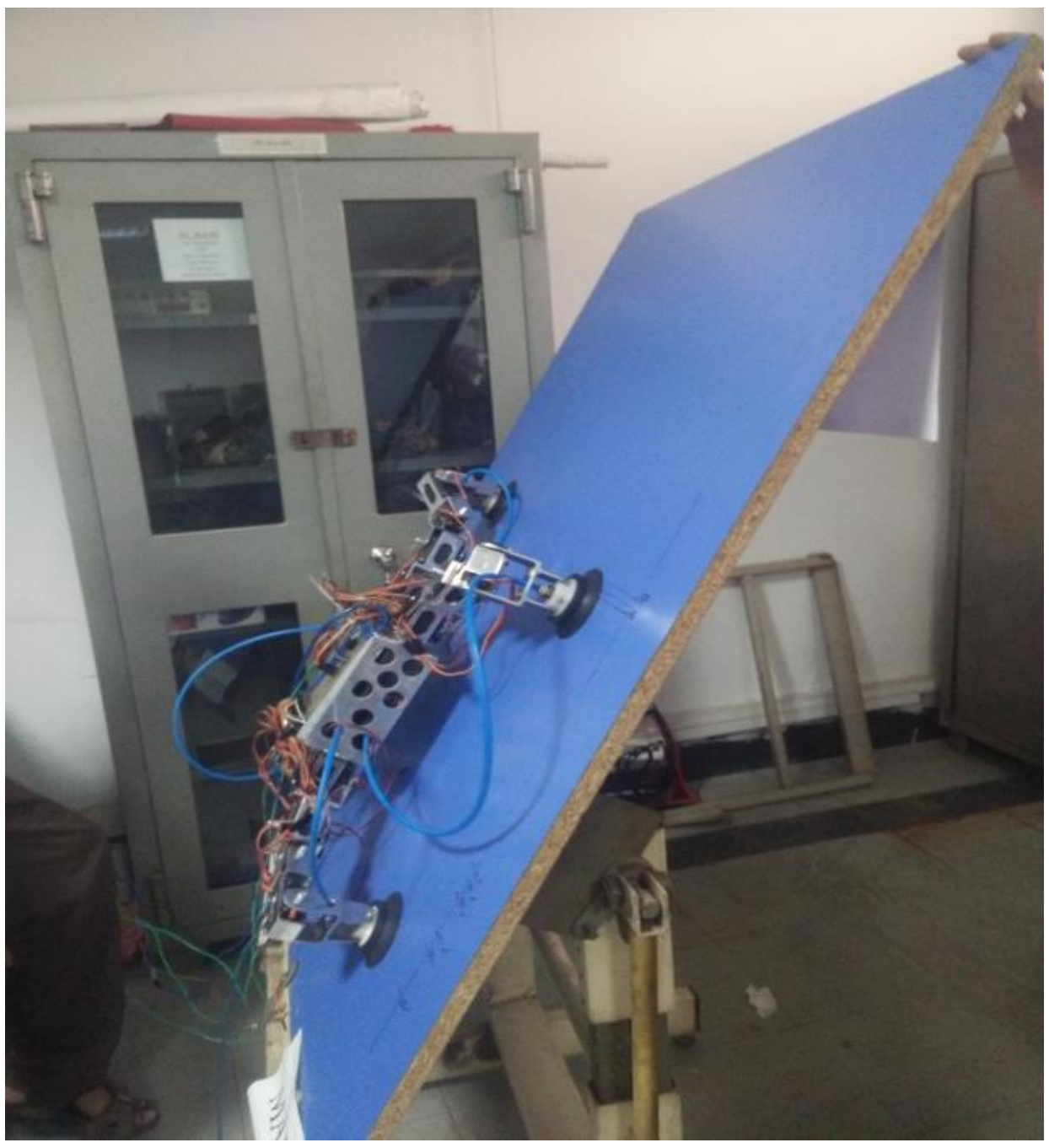

[Figure 16] Robot Testing

## Conclusion

The design and fabrication of the wall climbing robot has been successfully achieved as shown in figure [15-16]. The components used are easily available and programming required is simple. The designed robot has a stable structure and has shown a capability of climbing on vertical surfaces with a stable climbing gait. The robot has the ability to be used with other adhesion techniques as well by just replacing the end effectors. However, the robot structure is made of aluminium and on board battery pack and suction pumps are used which makes it quite heavier. Light weight batteries and light weight fibre made structure should be used to reduce the weight for better climbing performance. End effector design using different adhesion methods should be designed, manufactured and tested. Different path planning algorithms can be implemented and tested. Finally, this robot should be further developed by incorporating environment awareness using different sensors and autonomous climbing capability can be achieved to perform tasks at difficult to reach places where human access is difficult.

## Acknowledgements

This work was done as final year project of bachelor degree at University of Engineering and Technology, UET, Lahore. Thanks to Robotics Lab staff for their support and Thanks to Mechatronics department UET Lahore for funding.